\documentclass{article}

\usepackage{colm2024_conference}

\usepackage{booktabs}
\usepackage{graphicx}
\usepackage{placeins}
\usepackage[T1]{fontenc}
\usepackage[utf8]{inputenc}
\usepackage{amsmath}
\usepackage{amssymb}
\usepackage{amsfonts}
\usepackage{nicefrac}

\newcommand{\matS}{\mathbf{S}}
\newcommand{\matI}{\mathbf{I}}
\newcommand{\vk}{\mathbf{k}}
\newcommand{\ve}{\mathbf{e}}
\newcommand{\vv}{\mathbf{v}}
\newcommand{\vq}{\mathbf{q}}
\newcommand{\vzero}{\mathbf{0}}
\newcommand{\matM}{\mathbf{M}}
\newcommand{\matD}{\mathbf{D}}
\newcommand{\matP}{\mathbf{P}}
\newcommand{\matH}{\mathbf{H}}
\newcommand{\matA}{\mathbf{A}}
\newcommand{\matK}{\mathbf{K}}
\newcommand{\matV}{\mathbf{V}}
\newcommand{\matQ}{\mathbf{Q}}
\newcommand{\matO}{\mathbf{O}}
\newcommand{\matW}{\mathbf{W}}
\newcommand{\matU}{\mathbf{U}}
\newcommand{\vw}{\mathbf{w}}
\newcommand{\vu}{\mathbf{u}}

\author{\fontsize{9.3}{11.5}\selectfont
\textbf{Xiao Li}$^{1,2}$,
\textbf{Chengruidong Zhang}$^{1}$,
\textbf{Hao Luo}$^{1}$,
\textbf{Xi Lin}$^{1,3}$,
\textbf{Zekun Wang}$^{1}$,
\textbf{Zihan Qiu}$^{1}$,
\textbf{Yunfei Mao}$^{1}$,
\textbf{Langshi Chen}$^{1}$,
\textbf{Man Yuan}$^{1}$,
\textbf{Minmin Sun}$^{1}$,
\textbf{Huiqiang Jiang}$^{1}$,
\textbf{Siqi Zhang}$^{1}$,
\textbf{Rui Men}$^{1}$,
\textbf{Wei Hu}$^{2}$,
\textbf{Gong Cheng}$^{2}$,
\textbf{Bo Zheng}$^{1\dagger}$,
\textbf{Dayiheng Liu}$^{1\dagger}$,
\textbf{Jingren Zhou}$^{1}$\\
$^1$Qwen Team\quad $^2$Nanjing University\quad $^3$Zhejiang University\\
\small $^\dagger$Corresponding authors.
}

\title{Erase-then-Delta Attention:\\
Decoupling Erase and Write Addresses in Delta-Rule Linear Attention}

\begin{document}

\maketitle

\begin{abstract}
Delta-rule linear attention improves recurrent memory updates by correcting what is already stored at the current write address before writing new content. However, the active correction is still anchored to that same write address. As a result, stale information stored at a different address cannot be actively removed before new content is written elsewhere.
\quad
We propose \textbf{Erase-then-Delta Attention} (EDA), a memory update rule that decouples where to erase from where to write. The key insight is that recurrent memory models should not only correct the current write, but also selectively suppress outdated memory at an independently chosen address. Concretely, our method first applies a targeted erase step along a learned erase direction, and then performs the standard delta-style corrective write along the current write direction. This preserves the corrective behavior of delta-rule updates while expanding their memory-management capacity.
\quad
Language-model pretraining experiments across dense 2.5B and MoE 25B-A2.8B model families show that EDA performs best in both settings. The gain persists after 80B-token long-context midtraining of the MoE models, where EDA also performs best in long-context evaluations from 4k to 128k contexts. A compact update analysis and memory-state probes suggest why: EDA keeps the delta-rule corrective write intact while allocating an additional cleanup path most strongly when passive decay is weak. These results suggest that recurrent memory models should decide not only what to write, but also what stale information to erase and where.
\end{abstract}

\section{Introduction}
\label{sec:intro}

Autoregressive Transformers~\citep{vaswani2017attention} have become the foundation of modern language modeling, in part because softmax-based self-attention enables efficient parallel computation. This mechanism achieves strong performance on in-context learning and long-context retrieval by maintaining an explicit key--value cache. However, it also introduces fundamental bottlenecks at inference time: quadratic time complexity and linearly growing memory overhead that limit scalability for long-sequence tasks and agentic reasoning trajectories. To address these constraints, a growing body of work has explored efficient alternatives that maintain constant memory and $\mathcal{O}(1)$ inference time while preserving the expressive power of attention.

Recurrent models based on linear attention~\citep{katharopoulos2020transformers} and state space models~\citep{gu2021efficiently,gu2023mamba} offer a principled solution: they compress contextual information into a fixed-size state, enabling constant memory and linear-time training. Early variants such as Linformer~\citep{wang2020linformer} and RetNet~\citep{sun2023retentive} lacked data-dependent memory control and underperformed softmax attention. Subsequent models introduced dynamic gating mechanisms~\citep{yang2025gated,dao2024transformers,beck2024xlstm}, allowing selective forgetting and significantly narrowing the performance gap. However, additive gated updates still write new content into a finite state without explicitly correcting the association currently stored at the write address.

A more recent line of work replaces additive updates with the \emph{delta rule}~\citep{schlag2021linear}, which treats the recurrent state as a learnable associative memory that corrects itself toward the current key--value mapping. Gated DeltaNet (GDN)~\citep{yang2025gated} combines this corrective write with a head-wise forget gate, and recent channel-wise variants further refine this gate into a diagonal decay that gives each key feature its own retention rate~\citep{team2025kimi}. GDN-2 further separates the scalar delta gate into key-side erase and value-side write gates, but the active edit remains organized around the current write key~\citep{hatamizadeh2026gdn2}. 
We build on this channel-wise gated delta setting, also known as diagonal-plus-low-rank (DPLR), which combines GDN's hardware-efficient delta-rule structure with finer-grained channel-wise forgetting.
Despite this progress, a structural limitation remains unaddressed: the active delta correction still uses the current write direction $\vk_t$ as its only address. This coupling means the model can only suppress memory at the address it is currently writing to; stale information stored elsewhere must either persist or decay through channel-wise but address-agnostic forgetting.

This limitation has tangible consequences. In language modeling and state-tracking tasks, useful memory updates require not only writing new content but also removing obsolete information that would otherwise interfere with future reads and writes. When the model encounters a situation where earlier information must be invalidated---for example, a variable reassignment, a fact correction, or a context shift---it has no direct mechanism to remove the old content before committing the new one. The core missing capability is therefore not stronger forgetting, but \emph{targeted deletion of outdated memory at an address chosen independently of the current write}.

We address this problem with \textbf{Erase-then-Delta Attention} (EDA), a memory update rule that decouples erasure from writing. Instead of tying memory suppression to the current write address, EDA first removes stale content at an independently selected address and then performs the usual delta-style corrective write at the current write address. Intuitively, the erase step actively clears obsolete memory, while the delta step preserves the corrective writing behavior that makes delta-rule models effective. This yields a strictly richer update rule: the model can erase at one address and write at another within the same recurrent step.

We show that this simple modification has three important consequences. First, it provides a cleaner memory-management view of channel-wise gated delta recurrence by separating diagonal decay, independently addressed erasure, and write-coupled correction. Second, empirical analysis reveals that the model learns a near-orthogonal separation between erase and write addressing, indicating that the two operations serve genuinely different roles. Third, language-model pretraining experiments show that EDA improves over a DPLR-style gated delta baseline and compares favorably with several strong update-rule variants.

In summary, we introduce EDA, a gated delta-rule linear-attention update that decouples erase and write addresses while preserving the standard delta corrective write. We analyze the resulting erase-then-delta update and evaluate it through language-model pretraining, long-context evaluation, and memory-state probes, showing that the extra address acts as a conditional cleanup path rather than merely stronger forgetting.

\section{Preliminary}
\label{sec:preliminary}

We briefly introduce the recurrent memory notation and the channel-wise gated delta update most relevant to our method. The key point is that a diagonal forget gate already provides fine-grained decay, but the active correction and writing remain tied to the same address.

\subsection{Notation and Linear Associative Memory}

We consider a recurrent memory state $\matS_t \in \mathbb{R}^{d_k \times d_v}$ updated at each step $t$. The key $\vk_t \in \mathbb{R}^{d_k}$ serves as a write address, the value $\vv_t \in \mathbb{R}^{d_v}$ is the content to store, and the query $\vq_t \in \mathbb{R}^{d_k}$ reads from memory through $\matS_t^\top \vq_t \in \mathbb{R}^{d_v}$.

Standard linear attention updates memory additively:
\begin{equation}
\matS_t = \matS_{t-1} + \vk_t \vv_t^\top, \, \qquad \mathbf{o}_t = \matS_t^\top \vq_t.
\label{eq:linear_attn}
\end{equation}
This rule is efficient but does not explicitly decide what stale information to suppress.

\subsection{Coupled Erasure and Corrective Writing}

DeltaNet~\citep{schlag2021linear,yang2024parallelizing} replaces additive writing with a corrective update derived from the reconstruction loss
\begin{equation}
\mathcal{L}_t^{\mathrm{delta}}(\matS) = \frac{1}{2} \lVert \matS^\top \vk_t - \vv_t \rVert^2.
\label{eq:delta_loss}
\end{equation}
Taking a gradient step with learning rate $\beta_t$ gives
\begin{equation}
\matS_t = (\matI - \beta_t \vk_t \vk_t^\top)\matS_{t-1} + \beta_t \vk_t \vv_t^\top.
\label{eq:deltanet}
\end{equation}
Rather than simply accumulating $\vk_t \vv_t^\top$, DeltaNet first corrects what memory currently returns at address $\vk_t$ and then writes the new content at that same address.

Gated DeltaNet~(GDN)~\citep{yang2025gated} augments this rule with a head-wise scalar forget gate $\alpha_t \in (0,1)$:
\begin{equation}
\matS_t = \alpha_t(\matI - \beta_t \vk_t \vk_t^\top)\matS_{t-1} + \beta_t \vk_t \vv_t^\top.
\label{eq:gdn_prelim}
\end{equation}
Here $\alpha_t$ provides uniform decay within a head, while $(\matI - \beta_t \vk_t \vk_t^\top)$ provides address-specific correction. However, the erase-and-write behavior is still coupled: the same key $\vk_t$ determines both where memory is strongly modified and where new content is written. As a result, GDN can strongly suppress only the address it is currently writing to.

Following Kimi Delta Attention~(KDA)~\citep{team2025kimi}, we use the channel-wise version of this GDN design, replacing the head-wise scalar forget gate with a diagonal decay $\matD_t=\operatorname{Diag}(\boldsymbol{\alpha}_t)$:
\begin{equation}
\matS_t = (\matI - \beta_t \vk_t \vk_t^\top)\matD_t\matS_{t-1} + \beta_t \vk_t \vv_t^\top.
\label{eq:channel_delta_prelim}
\end{equation}
The diagonal gate gives each key channel its own retention rate and makes the transition compatible with a diagonal-plus-low-rank view. This improves how strongly different channels are preserved or decayed, but it does not change the addressing structure of the delta update itself: the corrective modification is still anchored to the current write key. Therefore, even with channel-wise gating, stale information stored at a different address cannot be explicitly erased before writing new content elsewhere.

GDN-2 addresses a closely related coupling by separating the scalar delta gate into key-side erase and value-side write gates~\citep{hatamizadeh2026gdn2}:
\begin{equation}
\matS_t = \left(\matI - \vk_t \widetilde{\ve}_t^\top\right)\matD_t\matS_{t-1} + \vk_t \boldsymbol{z}_t^\top,
\qquad
\widetilde{\ve}_t=\boldsymbol{b}_t \odot \vk_t,\quad \boldsymbol{z}_t=\boldsymbol{w}_t \odot \vv_t.
\label{eq:gdn2_prelim}
\end{equation}
This decouples the channel-wise erase and write strengths inside the delta residual. However, the erase/read direction $\widetilde{\ve}_t$ is still constructed from the current write key $\vk_t$, and the correction is still committed along $\vk_t$. Thus GDN-2 relaxes the gate-level coupling, while the address-level coupling between erasure and writing remains.

This coupling is the limitation we target. If stale information is stored at an address different from the current write address, the diagonal gate can decay feature channels but cannot selectively remove that stale association before writing elsewhere.

\subsection{Relation to Recent Delta-Style Variants}

Recent linear-recurrent models often improve performance by enriching the transition rule or embedding delta-style memory updates inside stronger architectures. DeltaProduct~\citep{siems2025deltaproduct} increases transition expressivity through multiple Householder-like factors per step, while RWKV-7~\citep{rwkv} and Comba~\citep{comba2025linear} adopt richer structured transition parameterizations. Recent hybrid architectures further demonstrate that strong designs built around expressive channel-wise gated delta components can be highly competitive with full attention~\citep{team2025kimi}.

Our goal is different. We do not primarily seek a globally richer transition; instead, we introduce a missing memory-management capability: erasing stale memory at one address before performing the standard delta-style corrective write at another. In that sense, our method is best viewed as orthogonal to transition-enrichment approaches and potentially compatible with stronger channel-wise gated delta backbones.

\section{Method}
\label{sec:method}

\subsection{Overview}
Our goal is to extend gated delta-rule linear attention with a missing memory-management capability: selectively deleting stale memory at an address different from the current write address. To do this, we revisit the DPLR-style update rule and identify a structural coupling between active correction and writing. We then introduce \textbf{Erase-then-Delta Attention} (EDA), a sequential update rule that adds an independently addressed erase step before the standard delta-style corrective write. This section first formalizes the limitation of the decay-gated delta baseline, then derives the new rule, and finally discusses its algebraic structure and stability properties.

\subsection{Erase-Write Coupling in Gated Delta Updates}
We consider a recurrent memory state $\matS_t$ updated by a gated delta rule with diagonal decay:
\begin{equation}
\matS_t = (\matI - \beta_t \vk_t \vk_t^\top)\matD_t\matS_{t-1} + \beta_t \vk_t \vv_t^\top,
\qquad \matD_t=\operatorname{Diag}(\boldsymbol{\alpha}_t).
\label{eq:channel_delta}
\end{equation}
Here $\matD_t$ is a diagonal decay matrix with retention factors $\boldsymbol{\alpha}_t$, $\beta_t$ controls the delta-style correction strength, $\vk_t$ is the current write direction, and $\vv_t$ is the value vector written into memory. This update is effective because it is not a naive additive write: after diagonal decay, the factor $(\matI - \beta_t \vk_t \vk_t^\top)$ corrects the memory response along the current write direction, and the additive term $\beta_t \vk_t \vv_t^\top$ writes the new content at the same address.

Equation~\eqref{eq:channel_delta} already contains both fine-grained decay and address-specific correction. The diagonal gate $\matD_t$ decides which key channels persist, while the rank-1 term $(\matI - \beta_t \vk_t \vk_t^\top)$ induces stronger correction along the current write direction. GDN-2 relaxes the scalar-gate version of this coupling by separating key-side erase and value-side write gates~\citep{hatamizadeh2026gdn2}. However, the active correction remains structurally coupled to writing: the edit is still constructed from the current write key, and the correction is still committed along $\vk_t$. Consequently, these updates can only strongly suppress memory through the address they are currently writing to.

This coupling is the core limitation we address. If stale information is stored at an address different from the current write direction, the model has no direct mechanism to remove it selectively before performing the current write. Instead, it must rely on the decay gate $\matD_t$, which is not tied to a specific stale address, or wait until future writes happen to revisit that address. Our central design question is therefore: can a delta-rule memory model erase at one address and write at another within the same recurrent step?

\subsection{Erase-then-Delta: Decoupled Erase-Write Addressing}
EDA decouples cleanup from writing by inserting an independently addressed erase operator before the standard delta write:
\begin{equation}
\matS_t =
(\matI - \beta_t \vk_t \vk_t^\top)
(\matI - \gamma_t \ve_t \ve_t^\top)
\matD_t\matS_{t-1}
+ \beta_t \vk_t \vv_t^\top .
\label{eq:ed}
\end{equation}
The factors in Eq.~\eqref{eq:ed} are applied from right to left. The diagonal decay $\matD_t$ first attenuates retained key coordinates, the erase factor $(\matI-\gamma_t\ve_t\ve_t^\top)$ contracts the decayed memory along a learned cleanup address $\ve_t$, and the usual delta factor then performs corrective forgetting and writing at the current write key $\vk_t$. This order is part of the update rule: for diagonal decay, $\matD_t$ generally does not commute with the rank-1 erase operator unless $\matD_t$ degenerates to a scalar decay or $\ve_t$ lies in an equal-decay subspace.

To see what the new operator actually erases, let
\begin{equation}
\widehat{\matS}_t = \matD_t\matS_{t-1}
\label{eq:decayed_memory}
\end{equation}
denote the memory after diagonal decay and before address-selective cleanup. EDA defines the erase address through the online objective
\begin{equation}
\mathcal{L}^{\mathrm{erase}}_t(\widehat{\matS}_t) =
\frac{1}{2}\lVert \widehat{\matS}_t^\top \ve_t \rVert^2.
\label{eq:erase_loss}
\end{equation}
This objective penalizes the content currently returned when the decayed memory is queried at $\ve_t$. A gradient step with learning rate $\gamma_t$ gives
\begin{equation}
\widetilde{\matS}_t =
(\matI-\gamma_t\ve_t\ve_t^\top)\widehat{\matS}_t ,
\label{eq:erase_step}
\end{equation}
where $\ve_t$ is L2-normalized. Thus $\ve_t$ is not merely an extra projection: it is the address whose current memory response is explicitly pushed toward zero.

This readout-level view clarifies why the new direction is more targeted than stronger decay. For any query direction $\vq$, the erased memory reads out as
\begin{equation}
\widetilde{\matS}_t^\top \vq
=
\widehat{\matS}_t^\top \vq
- \gamma_t(\vq^\top\ve_t)\widehat{\matS}_t^\top \ve_t .
\label{eq:erase_readout}
\end{equation}
When $\vq=\ve_t$, Eq.~\eqref{eq:erase_readout} suppresses the response at the erase address by a factor of $1-\gamma_t$. When $\vq$ is orthogonal to $\ve_t$, the erase step leaves that readout unchanged before the later delta update. The decay gate $\matD_t$ controls retention by key coordinate; in contrast, Eq.~\eqref{eq:erase_readout} subtracts the content currently returned at a learned memory address, scaled by how much the query aligns with that address.

After this cleanup, EDA applies the standard delta-style corrective write to the erased memory:
\begin{equation}
\matS_t =
(\matI-\beta_t\vk_t\vk_t^\top)\widetilde{\matS}_t
+ \beta_t\vk_t\vv_t^\top .
\label{eq:eda_two_stage}
\end{equation}
Substituting Eq.~\eqref{eq:erase_step} into Eq.~\eqref{eq:eda_two_stage} recovers Eq.~\eqref{eq:ed}. The delta correction and write at $\vk_t$ are therefore unchanged; the new degree of freedom is that stale memory can be suppressed at $\ve_t$ before new content is written at $\vk_t$. If $\ve_t$ collapses to $\vk_t$, EDA reduces to a stronger same-address correction; when the two directions differ, cleanup and writing are no longer forced to use the same address. This also distinguishes EDA from gate-level erase/write separation, where the residual can be reweighted by gates but remains organized around the current write key.

\begin{figure}[!ht]
\centering
\includegraphics[width=0.7\linewidth]{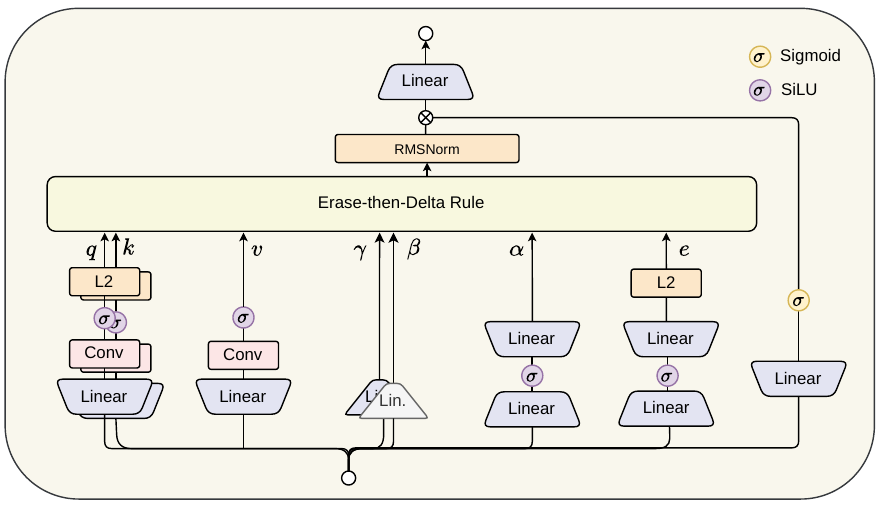}
\caption{Architecture of an EDA layer. The input is projected into query, key, value, output gate, erase gate~($\gamma$), delta gate~($\beta$), decay parameters~($\alpha$), and erase address~($\ve$). The query, key, and erase address are L2-normalized; the erase address uses a low-rank projection. All signals feed into the EDA kernel, whose output is normalized and gated before a final linear projection.}
\label{fig:eda_arch}
\end{figure}

The resulting rule separates memory management into three levels of specificity: diagonal decay through $\matD_t$, independent directional erasure through $\gamma_t\ve_t$, and write-coupled correction through $\beta_t\vk_t$. In this sense, EDA adds the missing degree of freedom needed to suppress stale memory at one address before performing a corrective write at another. 
Figure~\ref{fig:eda_arch} illustrates the full EDA layer architecture.

\paragraph{Safe gate for bounded decay.}
The diagonal decay in Eq.~\eqref{eq:ed} is parameterized in log space. Let $\matD_t=\operatorname{Diag}(\exp(\boldsymbol{g}_t))$, $\mathbf{A}=\exp(\mathbf{A}_{\log})>0$, $\boldsymbol{u}_t=\boldsymbol{a}_t+\boldsymbol{b}_{\Delta}$, and $\boldsymbol{\Delta}_t=\operatorname{softplus}(\boldsymbol{u}_t)$, where $\boldsymbol{a}_t$ is the decay projection and $\boldsymbol{b}_{\Delta}$ is a learned bias. The Mamba2/GDN-style log-space gate~\citep{dao2024transformers,yang2025gated} uses
\begin{equation}
\boldsymbol{g}_t^{\mathrm{log}}
= -\mathbf{A}\odot \boldsymbol{\Delta}_t,
\label{eq:log_gate}
\end{equation}
which guarantees $\exp(\boldsymbol{g}_t)\leq 1$ but leaves the log-decay unbounded below. KDA computes its safe gate as $\boldsymbol{g}_t^{\mathrm{KDA}}=\ell\,\sigma(\mathbf{A}\odot\boldsymbol{u}_t)$ with $\ell<0$, mapping each log-decay coordinate into $(\ell,0)$~\citep{team2025kimi}. EDA instead uses a bounded safe gate with the same lower log-decay limit and maximum value $0$:
\begin{equation}
\boldsymbol{g}_t
= \ell + (-\ell)\exp\!\left(
-\frac{\mathbf{A}}{\lvert\ell\rvert}\odot
\boldsymbol{\Delta}_t
\right),
\label{eq:safe_gate}
\end{equation}
where the exponential is applied elementwise. Since $\exp(-x)\in(0,1]$ for $x\geq0$, this parameterization keeps $\boldsymbol{g}_t\in(\ell,0]$ and therefore bounds each decay coordinate by $\exp(\ell)<\alpha_{t,i}\leq1$.

Comparing Eq.~\eqref{eq:log_gate} and Eq.~\eqref{eq:safe_gate} shows why this bounded form is useful beyond numerical clipping. The Mamba2/GDN-style log-space gate separates two roles: $\mathbf{A}$ controls the decay magnitude, while $\boldsymbol{\Delta}_t=\operatorname{softplus}(\boldsymbol{u}_t)$ acts as a ReLU-like nonnegative switch for whether a coordinate should decay. Our safe gate preserves this amplitude--switch decomposition near the active region: elementwise, a Taylor expansion around $\Delta_{t,i}=0$ gives $g_{t,i}=-A_i\Delta_{t,i}+\mathcal{O}(A_i^2\Delta_{t,i}^2/\lvert\ell\rvert)$. It therefore behaves like the log-space gate for small decay inputs, but saturates for large inputs instead of driving the log-decay toward $-\infty$. By contrast, the KDA sigmoid gate bounds the log-decay by applying a sigmoid directly to the affine decay signal, so $\mathbf{A}$ mainly changes the sigmoid slope and saturation rather than acting as a separate decay-amplitude parameter. In practice we set $\ell=-5$, making the smallest per-step decay factor $\exp(\ell)\approx6.7\times10^{-3}$, well within the normal range of half-precision formats. This prevents decay factors from becoming subnormal or zero, allowing decay-weighted chunk tensors to remain in half precision and preserving Tensor-Core-friendly dense matrix multiplications.

\paragraph{Cross-term structure and update order.}
The order in Eq.~\eqref{eq:ed}---erase first, then delta---is essential. Expanding the product of the two rank-1 operators reveals why:
\begin{equation}
(\matI - \beta_t \vk_t \vk_t^\top)(\matI - \gamma_t \ve_t \ve_t^\top)
= \matI
- \gamma_t \ve_t \ve_t^\top
- \beta_t \vk_t \vk_t^\top
+ \gamma_t \beta_t (\vk_t^\top \ve_t)\, \vk_t \ve_t^\top.
\label{eq:expansion}
\end{equation}
The final term---the cross-term---is proportional to the cosine similarity $c_t = \ve_t^\top \vk_t$ between the erase and write directions. It quantifies the ``leakage'' that occurs when the two directions are not orthogonal: the erase operation can influence the subsequent write-address correction through $\vk_t \ve_t^\top$. Reversing the order (delta first, then erase) would apply the erase operator after the write, allowing memory cleanup to suppress newly written content. By applying erasure first, our rule ensures that cleanup acts on old content before the new corrective write is committed.

When the model learns a near-orthogonal separation between $\ve_t$ and $\vk_t$ (mean $|c_t| \approx 0.105$, see Figure~\ref{fig:eda_memory_state_analysis}(c)), the cross-term becomes small and the update is well-approximated by two independent corrections acting on orthogonal subspaces. In this regime, the sequential rule is stable and the first-order effects dominate.

\subsection{EDA with Chunk-wise Parallel}
Referring to Eq.~\eqref{eq:ed}, the EDA state is multiplied by two rank-1 correction factors per step. To reuse existing DPLR chunk-wise kernels, we interleave the erase and delta sub-steps into a doubled sequence of length $2t$. Let

\begin{equation}
[\vq'_\tau, \vk'_\tau, \vv'_\tau, \beta'_\tau, \boldsymbol{\alpha}'_\tau] = \left\{
\begin{aligned}
&[\vzero, \ve_t, \vzero, \gamma_t, \boldsymbol{\alpha}_t], &\tau&=2t-1 \\
&[\vq_t, \vk_t, \vv_t, \beta_t, \mathbf{1}], &\tau&=2t
\end{aligned}\right.
\label{eq:double_data}
\end{equation}

Each original step $t$ maps to two sub-steps in the doubled sequence: the odd sub-step applies the erase operator with decay, and the even sub-step applies the delta correction with identity decay. This reduces EDA to a standard DPLR recurrence over twice as many steps. We can rewrite Eq.~\eqref{eq:ed} as:

\begin{equation}
\begin{aligned}
\matS_t &=
(\matI - \beta_t \vk_t \vk_t^\top)
(\matI - \gamma_t \ve_t \ve_t^\top)
\matD_t\matS_{t-1}
+ \beta_t \vk_t \vv_t^\top \\
&= (\matI - \beta_t \vk_t \vk_t^\top) \matI
\left( (\matI - \gamma_t \ve_t \ve_t^\top) \matD_t\matS_{t-1} + 0 \right)
+ \beta_t \vk_t \vv_t^\top \\
&= (\matI - \beta'_{2t} \vk'_{2t} \vk_{2t}'^\top) \matD'_{2t}
\left( (\matI - \beta'_{2t-1} \vk'_{2t-1} \vk_{2t-1}'^\top) \matD'_{2t-1}\matS_{t-1} + \beta'_{2t-1}\vk'_{2t-1} \vv_{2t-1}'^\top \right)
+ \beta'_{2t} \vk'_{2t} \vv_{2t}'^\top
\end{aligned}
\label{eq:double_kda}
\end{equation}

By partially expanding the recurrence for Eq.~\eqref{eq:double_kda} into a chunk-wise formulation, we have:

\begin{equation}
\matS_t = \underbrace{\left( \prod_{i=1}^{2t} \left( \matI - \beta'_i \vk'_i \vk_i'^\intercal \right) \matD'_i \right)}_{:=\matP} \matS_0
+ \underbrace{\sum_{i=1}^{2t} \left( \prod_{j=i+1}^{2t} \left( \matI - \beta'_j \vk'_j \vk_j'^\intercal \right)  \matD'_j \right) \beta'_i \vk'_i \vv_i'^\intercal}_{:=\matH}
\label{eq:chunk}
\end{equation}

Following the chunk-wise algorithm of KDA~\citep{team2025kimi}, we apply \textbf{WY representation} to pack a series of updates into a single compact representation:

\begin{equation}
\begin{aligned}
\vw_{2t} &= \beta'_{2t} \left( \underbrace{\left(\prod_{i=1}^{2t}\matD'_i\right)}_{:= \matD'_{1 \to 2t}} \vk'_{2t} - \sum_{i=1}^{2t-1} \vw_i \left( \vk_i'^\intercal \underbrace{\left(\prod_{j=i}^{2t}\matD'_j\right)}_{:=\matD'_{i \to 2t}} \vk'_{2t} \right) \right) \\
\vu_{2t} &= \beta'_{2t} \left( \vv'_{2t} - \sum_{i=1}^{2t-1} \vu_i \left( \vk_i'^\intercal \matD'_{i \to 2t} \vk'_{2t} \right) \right) \\
\matP &= \matD'_{1 \to 2t} - \sum_{i=1}^{2t} \matD'_{i \to 2t} \vk'_{i} \vw_i^\intercal \\
\matH &= \sum_{i=1}^{2t} \matD'_{i \to 2t} \vk'_{i} \vu_i^\intercal
\end{aligned}
\label{eq:wy_representation}
\end{equation}

And \textbf{UT transform} to reduce non-matmul FLOPs:

\begin{equation}
\begin{aligned}
\matA_{1 \to 2t} &= \left[ \begin{array}{c|c|c|c}
\mathrm{diag}(\matD'_{1 \to 1}) & \mathrm{diag}(\matD'_{1 \to 2}) & \cdots & \mathrm{diag}(\matD'_{1 \to 2t})
\end{array} \right] \\
\matA_{i \to 2t} &= \left[ \begin{array}{c|c|c|c}
\mathrm{diag}(\matD'_{1 \to 2t}) & \mathrm{diag}(\matD'_{2 \to 2t}) & \cdots & \mathrm{diag}(\matD'_{2t \to 2t})
\end{array} \right] \\
\matM &= \left( \matI + \mathrm{StrictTril}\left( \mathrm{Diag}(\beta') \left( \matA_{1 \to 2t} \odot \matK' \right) \left( \frac{\matK'}{\matA_{1 \to 2t}} \right)^\intercal \right) \right) \\
\matW &= \matM \left( \matA_{1 \to 2t} \odot \matK' \right) \\
\matU &= \matM \matV' \\
\end{aligned}
\label{eq:ut_transformation}
\end{equation}

Finally, the state and output can be computed in a chunk-wise manner using the matrix form:

\begin{equation}
\begin{aligned}
\matS_t &= \matD'_{1 \to 2t} \matS_0 + \left( \matA_{i \to 2t} \odot \matK' \right)^\intercal (\matU - \matW \matS_0) \\
\matO &= \left( \matA_{1 \to 2t} \odot \matQ' \right) \matS_0 + \mathrm{Tril}\left( \left( \matA_{1 \to 2t} \odot \matQ' \right) \left( \frac{\matK'}{\matA_{1 \to 2t}} \right)^\intercal \right)(\matU - \matW \matS_0)
\end{aligned}
\label{eq:chunk_wise_eda}
\end{equation}

This formulation reduces EDA's two-factor update to the standard DPLR chunk-wise recurrence.

\subsection{Efficiency Analysis}

The chunk-wise parallel formulation above increases the per-chunk sequence length, which raises the compute workload during prefill. However, the only additional inputs to the kernel are the erase address $\ve$ and the scalar gate $\gamma$, so the increase in HBM traffic remains modest after kernel fusion. Since the chunk-forward pass of channel-wise gated delta models is inherently memory-bound, the wall-clock overhead remains moderate in practice. During autoregressive decoding the effect is smaller still, as the dominant cost is reading and writing the recurrent state rather than computing the rank-1 updates. Moreover, linear-attention layers typically account for a minor fraction of end-to-end model latency, further limiting the overall impact. Optimized kernel implementations will be released at \url{https://github.com/QwenLM/FlashQLA}.

\section{Experiments}
\label{sec:experiments}

\subsection{Experimental Setup}

We evaluate EDA under two matched pretraining scales: a dense 2.5B model family and a larger MoE 25B-A2.8B family. The goal is to test whether the proposed erase-then-delta update improves the recurrent component in both a standard dense setting and a sparse-activated large-model setting. Within each scale, the compared models share the same training setup; detailed architecture hyperparameters and parameter counts are listed in Appendix~\ref{app:model_configs}.

\paragraph{Compared models.}
For the dense comparison, we compare a full-attention Transformer baseline with GDN, GDN-2, KDA, and EDA. For the MoE comparison, we compare GDN, KDA, and EDA under the same sparse-activation backbone. Except for the Transformer baseline, all compared linear attention models are hybrid architectures with three linear-attention layers followed by one full-attention Transformer layer, corresponding to a 3:1 linear-to-full attention ratio. This ratio is not tuned specifically for EDA; it follows the common hybrid configuration used in Qwen3.5-style and Kimi Linear architectures~\citep{qwen3next2025,team2025kimi}.

\paragraph{Training setup.}
All models were pretrained for 400B tokens with sequence length 4096 and global batch size 1024. The dense models used a learning rate decayed from $4\times10^{-3}$ to $3\times10^{-5}$, while the MoE 25B-A2.8B models used a learning rate decayed from $2\times10^{-3}$ to $3\times10^{-5}$. We additionally report MoE checkpoints after an 80B-token midtraining stage initialized from the 400B-token pretrained MoE checkpoints. The midtraining stage used sequence length 32k.

\paragraph{Evaluation setup.}
For downstream evaluation, we report MMLU, MMLU-Pro, GSM8K, MATH, BBH, and EvalPlus. Unless otherwise stated, entries are percentages averaged over two evaluation runs of the same checkpoint, and the Avg. column denotes the unweighted mean over the displayed benchmarks. Brief descriptions of the downstream benchmarks are provided in Appendix~\ref{app:evaluation_benchmarks}.

\subsection{Model Results}

\begin{table}[t]
\centering
\caption{Evaluation results after 400B-token pretraining. Values are percentages averaged over two evaluation runs; Avg. is the unweighted mean over the six benchmark columns. Within each model family, best results are bold and second-best results are underlined.}
\label{tab:model_results}
\small
\setlength{\tabcolsep}{3.8pt}
\begin{tabular}{lrrrrrrr}
\toprule
Model & MMLU & MMLU-Pro & GSM8K & MATH & BBH & EvalPlus & Avg. \\
\midrule
\multicolumn{8}{l}{\textit{Dense 2.5B}} \\
Transformer & \textbf{50.11} & \textbf{18.11} & 20.26 & 12.16 & 35.01 & 30.82 & 27.75 \\
GDN & 49.99 & 15.76 & 20.83 & 12.85 & 34.99 & 31.08 & 27.58 \\
GDN-2 & 49.90 & 15.84 & \underline{21.40} & \textbf{13.56} & 35.18 & \textbf{32.95} & \underline{28.14} \\
KDA & \underline{50.03} & \underline{16.37} & 21.30 & 13.09 & \underline{35.31} & 30.73 & 27.81 \\
EDA & 49.27 & 16.04 & \textbf{23.90} & \underline{13.54} & \textbf{35.42} & \underline{32.50} & \textbf{28.44} \\
\addlinespace[0.4em]
\midrule
\multicolumn{8}{l}{\textit{MoE 25B-A2.8B}} \\
GDN & \underline{64.75} & 31.46 & \textbf{58.89} & \underline{31.87} & \underline{51.88} & 50.09 & 48.16 \\
KDA & 63.84 & \underline{33.14} & \underline{58.61} & 30.78 & 51.87 & \underline{51.60} & \underline{48.31} \\
EDA & \textbf{65.31} & \textbf{33.61} & 57.71 & \textbf{33.57} & \textbf{52.72} & \textbf{53.37} & \textbf{49.38} \\
\bottomrule
\end{tabular}
\end{table}

At the dense 2.5B scale, EDA achieves the strongest average score among all dense models. Compared with KDA, which shares the same channel-wise gated delta backbone but lacks the independent erase address, EDA improves the Avg.\ score by 0.63 points.

The larger MoE 25B-A2.8B setting gives a clearer picture of the scaling behavior. EDA performs best on most benchmarks and improves the overall evaluation performance across knowledge-heavy, reasoning-heavy, and code-oriented tasks. This larger-scale result suggests that address-level erase/write decoupling provides a broadly useful memory-management degree of freedom: the model can preserve the delta-rule correction at the current write key while using a separate learned address to suppress stale content elsewhere.

\subsection{Midtraining Results}

Table~\ref{tab:midtrain_results} reports the same benchmark suite after the MoE 25B-A2.8B checkpoints were further trained for 80B tokens at 32k sequence length.

\begin{table}[t]
\centering
\caption{Evaluation results for MoE 25B-A2.8B checkpoints after 400B-token pretraining followed by 80B-token midtraining at 32k sequence length. Values are percentages averaged over two evaluation runs; Avg. is the unweighted mean over the six benchmark columns. Best results are bold and second-best results are underlined.}
\label{tab:midtrain_results}
\small
\setlength{\tabcolsep}{3.8pt}
\begin{tabular}{lrrrrrrr}
\toprule
Model & MMLU & MMLU-Pro & GSM8K & MATH & BBH & EvalPlus & Avg. \\
\midrule
GDN & \underline{67.43} & 40.55 & 75.93 & 45.94 & 64.25 & 50.09 & 57.37 \\
KDA & 67.32 & \underline{40.60} & \textbf{76.21} & \underline{46.87} & \underline{65.04} & \textbf{53.82} & \underline{58.31} \\
EDA & \textbf{68.12} & \textbf{41.71} & \underline{75.99} & \textbf{49.28} & \textbf{65.81} & \underline{51.78} & \textbf{58.45} \\
\bottomrule
\end{tabular}
\end{table}

Midtraining tests whether the pretraining-stage advantage survives a harder adaptation setting rather than only appearing at the original 4k training length. After the 80B-token long-context stage, EDA continues to provide the strongest overall performance, with especially clear gains on knowledge and reasoning benchmarks such as MMLU, MMLU-Pro, MATH, and BBH. This persistence is important because long-context midtraining changes the operating regime of the recurrent state: the model must maintain useful information over longer spans while still removing outdated content that can interfere with later reads.

Combined with the 400B-token pretraining results, the midtraining result strengthens the main conclusion: decoupling erase and write addresses remains useful after the model is further trained for longer contexts, suggesting that the erase path is compatible with, rather than fragile under, subsequent sequence-length adaptation.

\subsection{Long-Context Evaluation}

We evaluate the midtrained MoE checkpoints on the RULER task from 4k to 128k context length. Since midtraining used 32k sequences, the 64k and 128k settings evaluate length extrapolation beyond the training context. Table~\ref{tab:ruler_results} reports the RULER score at each context length, aggregated over all sub-tasks and four evaluation runs. EDA outperforms both GDN and KDA in the short-context regime from 4k to 16k, and remains close to the two baselines from 32k to 128k.

\begin{table}[t]
\centering
\caption{RULER~\citep{ruler} long-context results for MoE 25B-A2.8B checkpoints after 400B-token pretraining and 80B-token midtraining at 32k sequence length. Values are percentages averaged over four evaluation runs; 64k and 128k are length-extrapolation settings. Avg. is the unweighted mean over the six displayed context lengths. Best results are bold and second-best results are underlined.}
\label{tab:ruler_results}
\small
\setlength{\tabcolsep}{5pt}
\begin{tabular}{lrrrrrrr}
\toprule
Model & 4k & 8k & 16k & 32k & 64k & 128k & Avg. \\
\midrule
GDN & 92.40 & \underline{90.09} & \underline{87.28} & \textbf{82.15} & 67.62 & \textbf{45.16} & \underline{77.45} \\
KDA & \underline{93.33} & 89.29 & 85.13 & 79.66 & \textbf{72.00} & 42.70 & 77.02 \\
EDA & \textbf{93.84} & \textbf{90.88} & \textbf{87.55} & \underline{81.52} & \underline{71.90} & \underline{44.22} & \textbf{78.32} \\
\bottomrule
\end{tabular}
\end{table}
\subsection{Memory-State Analysis}
\label{sec:memory_state_analysis}

The benchmark gains above do not by themselves explain why an additional erase address helps, since the delta update already has two ways to reduce old content: diagonal decay $\matD_t=\operatorname{Diag}(\boldsymbol{\alpha}_t)$ and write-coupled correction $(\matI-\beta_t\vk_t\vk_t^\top)$. Throughout this subsection we analyze a fixed layer and attention head unless stated otherwise: $\vk_t\in\mathbb{R}^{d_k}$ is the L2-normalized write key at token $t$, $d_k$ is the per-head key dimension, $\matI\in\mathbb{R}^{d_k\times d_k}$ is the identity matrix, $\boldsymbol{\alpha}_t\in(0,1]^{d_k}$ is the per-channel retention vector, and $\beta_t\in(0,1)$ is the delta correction gate. We therefore ask a narrower mechanistic question: when the recurrent state must remove stale content, does the model use the new erase address in a way that cannot be explained by these two existing contraction paths alone?

We first measure a \emph{gate-strength allocation}, not exact removed state energy. Recall that the diagonal decay is parameterized in log space as $\matD_t=\operatorname{Diag}(\exp(\boldsymbol{g}_t))$, where $\boldsymbol{g}_t\in\mathbb{R}^{d_k}$ is the per-channel log-retention vector. Therefore $\boldsymbol{\alpha}_t=\exp(\boldsymbol{g}_t)$ is the per-channel retention factor applied before the erase and delta operators. For token $t$ and head $h$, let
\begin{equation}
\bar{\alpha}_{t,h}=\frac{1}{d_k}\sum_{j=1}^{d_k}\alpha_{t,h,j}
\end{equation}
be the mean retention factor of the diagonal decay within that head, where $\alpha_{t,h,j}$ is the $j$-th key-channel retention value and the sum averages over all $d_k$ key channels. Below we write $\alpha=\bar{\alpha}_{t,h}$ for compactness. This averaging deliberately collapses the diagonal operator to a scalar summary, so it should not be used to compare the full operator rank or total energy removed by $\matD_t$ against the two rank-1 contractions. Its purpose is narrower: under the average retained scale of a head, we ask how the learned gates allocate contraction strength among the decay path, the write-key correction, and the independent erase path. Since both rank-1 operators act after $\matD_t$, we define the unnormalized scores
\begin{equation}
b_D = 1-\alpha,\qquad
b_\Delta = \alpha\beta_t,\qquad
b_E = \alpha\gamma_t,
\end{equation}
for diagonal decay, same-address correction, and independent erase, respectively; here $\gamma_t\in(0,1)$ is the erase gate. For readability, after fixing head $h$, we omit the head index on $\beta_{t,h}$ and $\gamma_{t,h}$ and write them as $\beta_t$ and $\gamma_t$. Here $1-\alpha$ is the average decay removal fraction obtained after summarizing the diagonal retention vector by its mean, rather than an exact operator-level decomposition of $\matD_t$. We plot the normalized share $q_m=b_m/(b_D+b_\Delta+b_E)$ for mechanism $m\in\{D,\Delta,E\}$. This definition is appropriate for the allocation question because $\beta_t$ and $\gamma_t$ are exactly the contraction factors of the two rank-1 readouts, while the multiplier $\alpha$ accounts for the fact that both contractions operate on the state retained after decay. It should not be read as an exact or fully fair state-energy decomposition: the actual content removed also depends on the anisotropic diagonal decay, the current state projections onto $\ve_t$ and $\vk_t$, and the overlap between the two addresses.

As a boundary check, we also evaluate raw write-key recall, which asks whether older hidden values can be read back from their original write keys. KDA performs better under this strict probe, so EDA's advantage should not be interpreted as uniformly better historical recall. This motivates focusing on cleanup allocation and erase-address structure rather than raw recall alone.

\begin{figure}[!h]
\centering
\includegraphics[width=\linewidth]{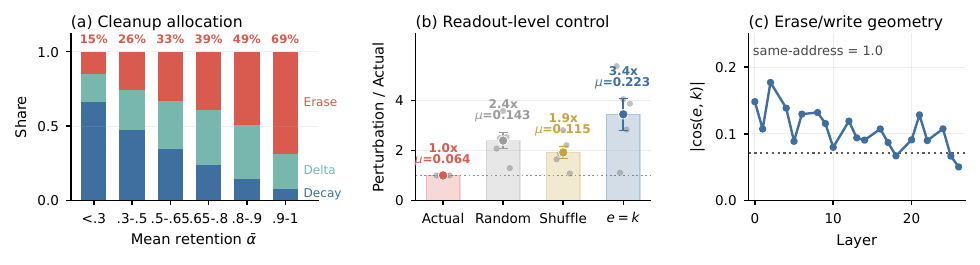}
\caption{\textbf{EDA uses an independent cleanup path.} (a) Gate-strength allocation by mean-retention bin. Independent erase becomes dominant when decay is weak ($\bar{\alpha}$ close to one); red percentages above bars denote the erase share. (b) Under the same erase gates, counterfactual erase directions cause larger local readout perturbations than the learned direction; bars show layerwise fold change relative to Actual, and $\mu$ denotes the raw mean perturbation score. (c) The erase address stays close to the independent-direction reference; same-address collapse would give $|\cos(e,k)|=1$.}
\label{fig:eda_memory_state_analysis}
\end{figure}

To test whether the learned erase direction is structured, we use two address-level diagnostics. First, we compare the readout-level control induced by the actual erase address $\ve_t\in\mathbb{R}^{d_k}$ with counterfactual directions: random unit vectors, head-shuffled learned erase directions, and the degenerate same-address choice $\ve_t=\vk_t$. For each direction strategy, we replay the recurrent state sequence with the same gates and measure the local effect of the erase step: $\mathbf{o}_t^{-}$ is the readout just before erase at token $t$, and $\delta \mathbf{o}_t$ is the readout change caused by that erase step. We compute the collateral perturbation score $\lVert\delta \mathbf{o}_t\rVert_2/\lVert\mathbf{o}_t^{-}\rVert_2$ over layers; the raw means are 0.064 for Actual, 0.143 for Random, 0.115 for Shuffle, and 0.223 for $\ve_t=\vk_t$. Figure~\ref{fig:eda_memory_state_analysis}(b) plots the same data as a layerwise fold change relative to Actual, with the raw means annotated. A smaller score does not mean ``no erase''; rather, under the same erase-gate budget, it means that the chosen address changes the currently readable state less than an alternative address. This probe therefore does not measure task benefit directly, but asks whether replacing the learned erase address causes larger collateral changes to the current readout. Second, we measure $|\cos(\ve_t,\vk_t)|$, the absolute cosine similarity between the L2-normalized erase address and write key, on GSM8K few-shot prompts. The independent reference for this geometry check is the analytic mean of $|\mathbf{u}^\top\mathbf{r}|$ for two independent random unit directions $\mathbf{u},\mathbf{r}\in\mathbb{R}^{128}$, where 128 is the per-head key dimension in this model.

Figure~\ref{fig:eda_memory_state_analysis} shows where the new erase degree of freedom is used. The allocation in Figure~\ref{fig:eda_memory_state_analysis}(a) shows that diagonal decay through $\matD_t$, same-address correction through $(\matI-\beta_t\vk_t\vk_t^\top)$, and independent erase through $(\matI-\gamma_t\ve_t\ve_t^\top)$ account for 35.4\%, 31.8\%, and 32.8\% of the global share, respectively. 
More importantly, the allocation shifts with decay speed in the expected direction: when $\bar{\alpha}<0.3$, decay already supplies most of the contraction strength, while for nearly persistent heads with $\bar{\alpha}\geq0.9$, independent erase contributes 69.1\% and is about $3.0\times$ the same-address correction contribution. This high-retention regime is exactly where stale content would otherwise survive $\matD_t$, so the model assigns the extra cleanup budget to $\gamma_t\ve_t$ rather than forcing it through the current write key.

The learned erase direction is also controlled at the readout level rather than arbitrary. Figure~\ref{fig:eda_memory_state_analysis}(b) shows that replacing $\ve_t$ with random, shuffled, or same-address alternatives increases local readout perturbation by about $2.4\times$, $1.9\times$, and $3.4\times$, respectively, after normalizing each analyzed layer by its Actual score. Thus, the learned erase address is not just an additional direction for removing state; under the same erase gates, it changes the current readout less than alternative directions, suggesting a more controlled cleanup operation. Figure~\ref{fig:eda_memory_state_analysis}(c) provides a complementary geometry check: the observed mean $|\cos(\ve_t,\vk_t)|$ stays around 0.105 across layers, close to the independent-direction reference and far from the value near one expected under same-address collapse. Together with the raw-recall boundary check above, these probes support the address-decoupling interpretation of EDA while clarifying its limitation: independent erase is a conditional cleanup mechanism, not a uniformly better historical-recall mechanism.

\section{Related Work}
\label{sec:related}

\paragraph{Delta-rule and gated linear memory models.}
DeltaNet~\citep{schlag2021linear,yang2024parallelizing} reinterprets recurrent state updates as online gradient descent on a reconstruction loss, replacing naive additive writes with corrective writes that depend on what is already stored at the current address. Gated DeltaNet (GDN)~\citep{yang2025gated} extends this with a head-wise forget gate, and recent channel-wise gated variants further replace that head-wise gate with diagonal decay for finer retention control~\citep{team2025kimi}. GDN-2 is the closest motivation-side comparison: it also argues that erase and write should be decoupled in delta-rule memory, but it targets a different axis of coupling~\citep{hatamizadeh2026gdn2}. Specifically, GDN-2 separates key-side erase and value-side write gates, allowing the model to assign different strengths to erasing and writing inside the delta residual. The active edit, however, remains organized around the current write key. EDA targets the complementary address-level coupling: it keeps the corrective delta write at $\vk_t$ while adding an independently addressed erase direction before the write. The two designs are therefore orthogonal in spirit: GDN-2 decouples how strongly erase and write are applied, while EDA decouples where erasing and writing are applied.

\paragraph{Expressive state-transition mechanisms for linear RNNs.}
A growing body of work seeks to enrich the state transition in linear RNNs beyond the single-step delta correction. DeltaProduct~\citep{siems2025deltaproduct} applies a sequence of Householder reflections per step, enabling smooth interpolation between diagonal and dense transitions. RWKV-7~\citep{rwkv} adopts a diagonal-plus-low-rank (DPLR) parameterization with vector-valued gating, improving state-tracking capacity. Comba~\citep{comba2025linear} proposes a scalar-plus-low-rank (SPLR) form motivated by closed-loop control theory, adding output correction alongside state feedback. These approaches increase the expressive power of state evolution globally. Our method is complementary but different in purpose: rather than enriching the transition matrix, we introduce a specific memory-management capability---selectively deleting stale memory at one address before performing a corrective write at another---while preserving the delta-rule structure.

\paragraph{Hybrid architectures and inference efficiency.}
The computational bottleneck of softmax attention at inference time has motivated hybrid architectures that combine full attention with linear recurrent layers. Models such as Jamba~\citep{lenz2024jamba} and Nemotron~\citep{gu2026jetnemotron} interleave sparse full-attention layers with predominantly linear recurrent layers, achieving a practical trade-off between quality and efficiency. Recent channel-wise gated delta hybrids demonstrate that this design can match or exceed full-attention quality while reducing KV cache usage substantially~\citep{team2025kimi}. EDA is orthogonal to these architectural choices: it improves the recurrent component itself, making it a candidate drop-in replacement for channel-wise gated delta layers in hybrid designs.

\section{Conclusion}
\label{sec:conclusion}

We introduced Erase-then-Delta Attention, an address-level modification to delta-rule linear attention that separates where the model erases from where it writes. Instead of relying only on diagonal decay or same-address delta correction to remove stale content, EDA first applies a learned erase operation at an independent address and then performs the corrective delta write at the current write key. This keeps the core delta-rule update intact while giving the recurrent state a more direct way to clean up memory that is not aligned with the current write.

Across dense 2.5B and MoE 25B-A2.8B pretraining, EDA achieves the strongest average performance among the compared models, and the advantage persists after long-context midtraining of the MoE checkpoints. The memory-state analysis further supports the intended mechanism: the learned erase path is used most strongly when passive decay is weak, and counterfactual erase directions cause larger readout changes under the same erase gates. These results suggest that recurrent memory models benefit from deciding not only what to write, but also where stale information should be removed.

\paragraph{Limitations.}
Our work has several limitations. Introducing the independent erase step reduces raw write-key recall, so the erase path should be understood as a conditional cleanup mechanism rather than a uniform improvement to memory fidelity. Additionally, the current probes measure gate allocation and readout perturbation but do not directly trace individual erase events to specific downstream prediction improvements.

\bibliographystyle{colm2024_conference}
\bibliography{references}

\appendix

\section{Model Configurations}
\label{app:model_configs}

The model configurations used in the evaluation are summarized in Tables~\ref{tab:scale_configs} and~\ref{tab:param_configs}. All evaluated models use the same vocabulary size (248{,}320); pretraining used 4096-token sequences, and the MoE midtraining stage used 32k-token sequences. Training used bfloat16 with the AdamW optimizer, SiLU activations in the FFN/MoE blocks, and RMSNorm with $\epsilon=10^{-6}$. The hybrid models use one full-attention Transformer layer in every four layers, placed after three linear-attention layers. The full-attention layers in both the Transformer baseline and the hybrid models use Gated Attention~\citep{qiu2026gated}. For parameter alignment, the dense Transformer baseline uses 8/4/4 query/key/value heads in its full-attention layers.

\begin{table}[!ht]
\centering
\caption{Scale-level architecture hyperparameters. ``Layers'' reports total layers with linear/full-attention counts in parentheses. ``Attn/KV'' denotes the query and key/value head counts in hybrid full-attention layers. ``LA K/V'' denotes the number of key/value heads in the linear-attention layers, and ``LA dim'' denotes their per-head dimensions. For the MoE scale, the FFN/expert column expert width.}
\label{tab:scale_configs}
\scriptsize
\setlength{\tabcolsep}{3.4pt}
\begin{tabular}{lccccccc}
\toprule
Scale & Layers & $d_{\mathrm{model}}$ & Attn/KV & LA K/V & LA dim & FFN/expert & MoE routing \\
\midrule
Dense 2.5B & 24 (18/6) & 2048 & 8/2 & 8/16 & 128/128 & 7424--7488 & -- \\
MoE 25B-A2.8B & 28 (21/7) & 2048 & 16/2 & 16/32 & 128/128 & 512 & 256 experts, top-8 activated + 1 shared \\
\bottomrule
\end{tabular}
\end{table}

\begin{table}[!ht]
\centering
\caption{Total and active parameter counts for evaluated model variants. Dense models activate all parameters; MoE models report both total parameters and the parameters active per token.}
\label{tab:param_configs}
\small
\setlength{\tabcolsep}{6pt}
\begin{tabular}{llcc}
\toprule
Scale & Model & Total params & Active params \\
\midrule
Dense 2.5B & Transformer & 2.5052B & 2.5052B \\
Dense 2.5B & GDN-2 & 2.5353B & 2.5353B \\
Dense 2.5B & GDN & 2.5035B & 2.5035B \\
Dense 2.5B & KDA & 2.5218B & 2.5218B \\
Dense 2.5B & EDA & 2.5295B & 2.5295B \\
\addlinespace[0.3em]
MoE 25B-A2.8B & GDN & 24.5676B & 2.7236B \\
MoE 25B-A2.8B & KDA & 24.6324B & 2.7885B \\
MoE 25B-A2.8B & EDA & 24.6558B & 2.8119B \\
\bottomrule
\end{tabular}
\end{table}

For parameter efficiency, variants with channel-wise forget gates use rank-16~(per-head) low-rank projections for the gate generator. The EDA MoE configuration uses a rank-16~(per-head) erase-address projection and a safe gate with lower bound $-5$.

\section{Evaluation Benchmarks}
\label{app:evaluation_benchmarks}

We evaluate the pretrained checkpoints on a compact set of standard language-model benchmarks. MMLU measures broad multitask knowledge across academic and professional subjects \citep{mmlu}, while MMLU-Pro increases the difficulty with more challenging questions and larger answer sets \citep{mmlupro}. GSM8K evaluates grade-school mathematical reasoning with word problems \citep{gsm8k}, and MATH evaluates more advanced competition-style mathematical problem solving \citep{math}. BBH covers difficult reasoning tasks selected from BIG-Bench \citep{bbh}. EvalPlus evaluates code generation with stricter test cases beyond the original HumanEval/MBPP-style checks \citep{evalplus}.

\end{document}